# NLML--a Markup Language to Describe the Unlimited English Grammar

Jiyou Jia

Institute for Interdisciplinary Informatics,
University of Augsburg, Germany
jiyou.jia@student.uni-augsburg.de

**Abstract**: In this paper we present NLML (Natural Language Markup Language), a markup language to describe the syntactic and semantic structure of any grammatically correct English expression. At first the related works are analyzed to demonstrate the necessity of the NLML: simple form, easy management and direct storage. Then the description of the English grammar with NLML is introduced in details in three levels: sentences (with different complexities, voices, moods, and tenses), clause (relative clause and noun clause) and phrase (noun phrase, verb phrase, prepositional phrase, adjective phrase, adverb phrase and predicate phrase). At last the application fields of the NLML in NLP are shown with two typical examples: NLOJM (Natural Language Object Modal in Java) and NLDB (Natural Language Database).

## 1   Introduction

Natural language processing (NLP) is a very important research field in artificial intelligence and computer linguistics. In the past 50 years great progress has been made in this field, as a lot of references have pointed out [1][7][9]. The milestone in the history of modern linguistics is Chomsky's concept of generative grammar for natural language with the efforts of describing all possible sentences [3][4][5][6]. Symbolic approaches to NLP have their origins in generative linguistics. Ideas from linguistic theory, particularly relative to syntactic description along the line proposed by Chomsky and others, were absorbed by researchers in the field, and refashioned in various ways to play a role in working computational systems; in many ways, research in linguistics and the philosophy of the language set the agenda for explorations in NLP.

But the actual English grammar is much more complicated and therefore a corresponding complicated rules system is needed to the English grammar. However work in syntactic description has always been the most thoroughly detailed and worked-out aspect of linguistic inquiry, so that at this level a great deal has been borrowed by NLP researchers. There has been many successful projects on syntactic formalisms and parsing algorithms based on the notation of Chomsky-grammar, like AGFL-project (Affix Grammars over a Finite Lattice) from University of Nijmegen [12], FDG (Functional Dependency Grammar) language analysis system from University of Helsinki [14], Transition network grammars and parsers described by Woods [16][17], Chart-based parsers described by Kay [11] and Horn-clause-based parsers described and compared with transition network systems by Pereira and Warren [13]. So with these parsers and a meticulous description of the grammar the sentences can be well decomposed into their grammatical elements.

However the problem is that only we, human being, know the meaning of the notations and their relations in the parsing results such as NP (Noun Part), VP (Verb Part), subject, object, etc., but the computer program doesn't recognize them as the grammar elements in the sentences and therefore can do nothing further more with the parsing result in the sense of natural language understanding. For example the AGFL (available at http://www.cs.kun.nl/agfl under GNU) is a powerful and practical formalism for the syntactic analysis of natural languages.

The AGFL can generate parsers from grammars and lexica for a natural language. The default output of an AGFL-parser is a normal indented parse-tree, which can also be obtained in the form of a labeled bracket tree. For example, we have generated an English parser using AGFL and our grammar and lexica for English, which will be introduced later. The normal parse tree produced by the English parser for the simple sentence "*I come*" is shown in Fig. 1.

```
segment
    statement
        simple statement
            simple complete statement without it noun clause
                pre circumstances
                mid circumstances
                simple SVOC phrase
                    subject(sing, first)
                        noun phrase(sing, first, nom)
                            noun part(sing, first, nom)
                                personal pronoun(sing, first, nom)
                                    LEX_PERSPRON(sing, first, nom)
                                        PERSPRON(sing, first, nom)
                                            "I"
                    VOC phrase(sing, first)
                        simple VOC phrase(sing, first)
                            all VOC phrase(sing, first, present)
                                real all VOC phrase(sing, first, present)
                                verb group(present, intr, to|on|around|about|through|over|along|ahead|apart|away|back|forth|forward|off,none,sing, first)
                                    verb group without modal(present, intr, to|on|around|about|through|over|along|ahead|apart|away|back|forth|forward |off,
none, sing, first)
                                        mid circumstances
                                        verb form(present, intr, to|on|around|about|through|over|along|ahead|apart|away|back|forth|forward|off,none,sing, first)
                                            LEX_VERBI(none, intr, to|on|around|about|through|over|along|ahead|apart|away|back |forth|forward|off)
                                                VERBI(none, intr, to|on|around|about|through|over|along|ahead|apart|away |back|forth |forward|off)
                                                    "come"
                post circumstances
                mid circumstances
```

**Fig. 1.** The indented tree output

Obviously the parsing program produces a symbol list, whose meanings can only be understood by those who are good at English grammar. If the computer is further confronted with such a question about the grammar of this sentence as "what is the verb word of the sentence?" or "what is the subject of the sentence?", a new program should be written to analyze the tree structure in the parsing result to find a branch with the label "verb" or "subject" whose leafs are the answers to the questions. So a simple search algorithm for the data structure "tree" is able to deal with this parsing result from a simple sentence. But for the sentences with more than one verb or more than one subject, such as "*the book written by the famous professor interests the students very much*", or "*The book you give me today interests me very much.*", therefore there are also two corresponding leafs in the parsing tree for "verb" or for "subject". This makes it difficult for the simple search algorithm to decide which verb or which subject is the suitable answer to the questions above. In NLP the complicated sentences with relative clauses, noun clauses, etc. should also be equally considered as the simple sentences with only one subject and one verb phrase.

Fortunately the parsing result of AGFL can be the third form--the transduced form which is wholly determined by the transduction specified in the grammar written by a grammar expert and does not have to look like a parse tree. So using this technique we have been trying writing the English grammars and lexicons in such a format that the syntactic and semantic description of English sentences in the form of markup language is possible. We call this format NLML (Natural Language Markup Language). Still with the example above, the parsing result in the form of NLML for the sentence "*I come*" is:

<mood>statement</mood><complexity>simple</complexity><subject><noun><type>perspronoun</type>
<word>I</word><numb>sing</numb><pers>first</pers><case>nom</case></noun></subject><verb_phra
se><verb_type>verb</verb_type><tense>present</tense><numb>sing</numb><pers>first</pers>
<verb_word> come </verb_word><circum></circum></verb_phrase>

This format looks like simpler than the former two tree results. In order to answer the grammar questions about this sentence we need only tell the computer a new program which can get the content between the tag pair <subject> and </subject>, or between the <verb_word> and </verb_word>. Even confronted with the parsing result in this format for a more complicated sentence this new program can be easily extended to get the grammar elements and to construct the their objects in object oriented programming language such as Java, what will be shown later. With the easy control method of markup language, such as insert, delete and change, we can manage the NLML to realize the operations on the original sentences, such as negation, from statement to question and vice versa, etc.

Moreover, once we have got this parsing result from a given sentence, we can save it directly as a simple String in any database for later usages and don't have to parse the original sentence again. This makes the long—term storage of the syntax and semantics of an expression in natural language possible, which is the key point in the analysis of historical discourse or text.

## 2    Description of unlimited English Grammar with NLML

Our goal is to describe all grammatically correct English expressions with the markup language. So a textbook for English grammar is needed. We have selected the book from Chalker [2] and the book from Hanks & Grandison [8], which the basic notations of English grammar in this paper come from. Additionally some new notations are used in order to make the NLML more clear and understandable, although these notations may not be the most appropriate ones.

## 2.1   English Expressions

In English we should deal with all kinds of **sentence**s, **clause**s, **phrase**s and **word**s, i.e. all of them are the permissible expressions in English. Here are some examples of these expressions appeared in an actual dialog between two persons (A and B), and their grammar categories.

A: I will buy a book tomorrow.        –sentence (statement)
B: Why?                                –word (query adverb)
A: Because I have got some money. –subordinate sentence supplying information for the last sentence
B: Which book will you buy?           –sentence (question)
A: That one you have read.            –noun phrase with a relative clause modifying it
B: What a pity!                       -phrase (terse exclamation, the essence is a noun word: pity)
A: What about this book?              -sentence (question, the essence is a noun phrase: the booh)
B: Terrible!                          –word (adjective)
A: What?                              -word (query pronoun)
B:  How terrible that book is!        -sentence (full exclamation)
A: Please tell me why you say that!   -sentence (order)
       ……

    By hearing or reading these texts one person who has learned the English grammar should be able to recognize their grammatical categories, i.e. statement, question, etc., consciously or unconsciously. Similarly we should mark this category with a tag pair so that the computer can get this category by parsing. Here we use the tag pair "<mood>…</mood>" to label the expression category. The notation "mood" is suitable to describe whether a sentence is declarative, interrogative, imperative or exclamative, while it is not appropriate to distinguish a noun phrase from an adjective phrase. We just use it to label the expression categories in the English expressions.

    We describe all permissible expressions in a context--free rule system with the writing style of AGFL to produce our markup language:

```
expression:
    #1. e.g.  I will buy a book tomorrow.
    statement, ["."] /
        "<mood>statement</mood>",statement;
    #2 e.g.  Which book will you buy?
    question, ["?"] /
        "<mood>question</mood>",question;
    #3 e.g.Please read the book!
     ["please"], order, ["please"], ["!"] /
        "<mood>order</mood>",order;
                        #e.g.  Please tell me why you say that!
     #4 e.g.  How terrible that book is!
    full exclamation, ["!"]/
        "<mood>full exclamation</mood>", full exclamation!
    #5 e.g.  How about this book?
    "how", "about", object phrase, ["?"] /
        "<mood>about</mood>", object phrase;
    #6 e.g.  What about this book?
    "what", "about", object phrase, ["?"]/
        "<mood>about</mood>", object phrase;
     #7 e.g.  what a pity!
    "what", subject(NUMB, PERS), ["!"]/
        "<mood>what terse exclamation</mood>", subject;
    #8 e.g.  How terrible!
    "how", predicative adjective, ["!"]/
        "<mood>how terse exclamation</mood>", predicative adjective;
```

```
  #9 e.g.  What? Who?
  query NP(NUMB, CASE), ["?"]/
        "<mood>np</mood>",  query NP;
#10 e.g. Why?
 LEX_QUEADV(ATTRIBUTE)/
        "<mood>circumstances</mood>", LEX_QUEADV;
#11 e.g.  in the morning,  at home, certainly.
 $PENALTY(10), mid circumstances, ["."]/
        "<mood>circumstances</mood>",  mid circumstances;
 #12 e.g.  that book on the desk.
 $PENALTY(10), noun phrase(NUMB, PERS, CASE)/
        "<mood>np</mood>", noun phrase;
 #13 e.g.  That one you have read.
 $PENALTY(10), noun phrase(NUMB, PERS, CASE), relative clause(NUMB)/
        "<mood>np</mood>", noun phrase, relative clause;
 #14 e.g.  That one you have read.
 $PENALTY(10),  attribute adjectives /
        "<mood>adj</mood>",  attribute adjectives;
 #15 e.g.  ill.
 $PENALTY(10),  predicative adjective/
        "<mood>adj</mood>",  predicative adjective;
 #16 e.g.  Because I have got some money.
 $PENALTY(12),   subordinator, simple statement, [period]/
   "<mood>subcircum</mood><subordinator>", subordinator, "</subordinator>", simple statement.
```
(The line beginning with the symbol "#" is a commentary line so that we can use it
to write the explanation and examples for the rules and notation.)

With the notation above the parsing result of a given expression is at least one of
the 16 possibilities listed above, i.e. after the parsing the system can get the "mood" of
the expression from the tag pair "<mood>…</mood>". With the mood in mind the
system can continue its next parsing fit for every mood.

Whilst the other items end with semicolon, the 4th one ends with an interjection
("!"). That means the parsing result with the mood statement, question, order and full
exclamation, i.e. a full sentence with at least one subject and one verb phrase, has
greater priority over the others. We call this mechanism max-matching. In other
words if an expression can be recognized as a full sentence, there is no need to
continue to produce other parsing results any more. We do so because by reading or
hearing we try to recognize the text at first as a full sentence; if it is not a sentence, we
go on to deal with it as some kinds of phrases or words.

In the above listed 16 parsing results there are totally 11 categories labeled with the
tag "mood": "statement", "question", "order", "full exclamation", "np", "adj",
"about", "circumstances", "what terse exclamation", "how terse exclamation", and
"subcircum".

Those ones with the mood "statement", "question", "order", "full exclamation" and
"subcircum" all have at least one subject, therefore can be parsed next as a sentence.
In the following text we use the term "sentence" in this sense.

Those ones with the mood "np", "what terse exclamation", and "about" treat
essentially one noun phrase, therefore can be parsed next as a noun phrase.

Those ones with the mood "adj", and "how terse exclamation" treat essentially one
adjective phrase, therefore can be parsed next as an adjective phrase.

The one with the mood "circumstances" deals essentially with one circumstance
phrase therefore can be parsed next as a circumstance phrase.

In the following paragraphs we introduce the NLML of English grammar elements in the four levels: sentences, clauses, phrases and words. Limited by the paper length, we can only explain the construction of "sentences" more detailed and give the necessary NLML codes.

## 2.2   Sentences

In the sentence level there are several classification methods, for example, according to the mood, complexity, or voice.

### 2.2.1   Mood

Every sentence has its mood. So we use the tag pair <mood> and </mood> to label the mood of the sentence. The sentence moods and the corresponding examples are shown in table 1. We divide the exclamation sentences into three categories: what terse exclamation, how terse exclamation and full exclamation, as only the full exclamation can be treated as a sentence, and the terse exclamations should only be analyzed as noun phrase or adjective phrase.

### 2.2.2   Complexity

The sentences can also be classified according to their complexity. So we use the tag pair <complexity> and </complexity> to label the complexity of the sentences. The complexity types, some examples, and the NLML tags are shown in Table 2.

From the examples we can see that the compound complex sentence, compound sentence and complex sentence are made up with simple sentences connected by conjunctions and/or comma. Therefore the simple sentence is the most elementary type of all kinds of sentences.

The simple sentence is simpler compared with the compound complex, compound and complex sentences. But actually it may also contain very complicated sentences such as those with relative clause modifying the noun phrase and those with a noun clause.

### 2.2.3   Voice

Every sentence has its voice: active or passive. So the sentences can also be classified by their voice and we use the tag pair <voice> and </voice> to label the voice of the sentence. As there are just two types of voice, we treat the NLMLs without the tag pair <voice></voice> as with the active voice. The examples in table 2 are all in active voice. In table 3 there are examples with the passive voice. (The order sentence has no passive voice.)

**Table 1** Sentence types defined by their mood

| Mood | Example |
|---|---|
| Statement (declarative) | If it rains today, you can not go out, and I can not come. |
| Question (interrogative) | What will you do if it rains today? |
| Order (imperative) | Please do your homework if it rains today. |
| Full exclamation (exclamative) | What a rainy day it is! |

**Table 2** Sentences classified by their complexity with the active voice

| Complexity\mood | Statement | Question | Order | Exclamation |
|---|---|---|---|---|
| Compound complex | If it rains today, you will not go, and I will not come. | | If it rains today, please stay at home, listen to the radio and read the book! | |
| Compound | Today you come, he goes, and I wait.<br>It snows, but I still go out.<br>Neither you come, nor do I go. | What should I do, what can I do, and what must I do? | Please sit down, read the book and then write your paper!<br>Either live or die! | |
| Complex | If you come, I will go.<br>I lived whenever she lived. | What would he do if it rains today? | Please phone me if you have time. | |
| Simple | Both you and he come today.<br>Neither he nor I come today.<br>I don't understand what he is now saying.<br>I give him a book written by the famous professor.<br>I know the book you gave your girl friend yesterday.<br>The man coming today is my best friend.<br>The horse runs so fast that others can not catch up with it.<br>I see the student do his job carefully.<br>He has his car repaired. | Can you understand what he is saying?<br>Who is coming to fetch the book?<br>Whom did you give the book written by the famous professor? | Go to listen to the radio! | What a stupid man he is!<br>How beautiful she is! |

### 2.2.4   The relation between the Sentences with Different Complexities

As table 2 and 3 show, all other sentences can de decomposed into simple sentences connected by conjunctions. We demonstrate this in the order of the mood.

#### 2.2.4.1   *Statement*

A statement sentence is either a compound complex sentence, or a compound sentence, or a complex sentence, or a simple sentence. This can be described by the following notation:

```
statement:
    compound complex statement/
            "<complexity>compound complex</complexity>", compound complex statement!
    compound statement/
            "<complexity>compound</complexity>", compound statement!
    complex statement/
            "<complexity>complex</complexity>",  complex statement!
    simple statement/
            "<complexity>simple</complexity>", simple statement.
```

The max-matching mechanism is also used in parsing the statement sentence.

A complex statement sentence can either be a subordinate clause plus a main simple statement, or a main simple statement plus a subordinate clause. Between the subordinate clause and the main clause there may be, and sometimes must be, a comma. We use the tag pair <subordinator></subordinator> to label the type of the subordinate clause, and the content of this tag pair, except the "ever" and "whether or not", is the actual subordinator of the clause. We use the tag pair <sub> </sub> to label the simple sentence in the subordinate clause. In contrary there is no need to label the main clause, as we can parse the subordinate clause labeled with this tag pair at first and then the others as main clause.

The compound statement can be divided into two types: "and or statement", and "two simple statements". In "and or statement" the last independent simple statement is attached to other statement sentences with the conjunction "and" or "or", whilst the others are connected by comma. In "two simple statements" thee are just two simple statement sentences connected by such conjunctions as "but" or "neither…nor…".
We use the tag pair <simple_sentence> … </simple_sentence> to include one simple statement and <sentence_connector >…</sentence_connector> to point out the conjunction. If the conjunction is a single word like "so", "for", "but", or "yet", it is the actual conjunction connecting the two simple sentences. If it is two words connected by "-", like "either_or", or "neither_nor", it should be divided into two parts.

The "Compound complex statement" is actually a subordinate sentence plus an "and or statement" sentence.

Summarily the simple statement is the most elementary statement.

**Table 3.** Sentences classified by their complexity with the passive voice

| complexity\mood | statement | question | exclamation |
|---|---|---|---|
| compound complex | If it rains today, the desk should be moved into the room, and the window should be closed. | | |
| Compound | Today the car should be repaired, the room should be cleaned, and the clothes should be washed.<br>The car has been repaired, but the room has not been cleaned.<br>Neither the car is repaired, nor is the room cleaned. | What should be done, what can be done and what must be done? | |
| complex | If you come here, the room can be cleaned completely. | What should be done by us if it rains today? | |
| simple | Both the car and the bicycle are repaired by him alone.<br>Neither the car nor the bicycle was repaired by him.<br>What he is now saying can't be understood by me.<br>A book written by the famous professor is given him.<br>The student was seen to do his job carefully. | May the car be repaired by him?<br>Who was seen to do his job carefully?<br>How can the room be cleaned so completely? | What a good book has been lost by him!<br>How completely the room is cleaned! |

### 2.2.4.2   Simple Statement

As the AGFL doesn't support the left recursion nonterminals, we classify the simple statement into two types: "simple statement without noun clause as real subject" and "simple statement with noun clause as real subject". The first type has greater priority, so we give the second type a penalty. That means if an expression can be parsed both as the first type and as the second type, the first type will appear as the first choice.

The "simple statement without noun clause as real subject" is made up of "pre circumstances" and "simple SVerb phrase". The "pre circumstance" is the circumstance preceding the sentence. Some circumstances must be placed at the beginning of the sentence, some must be at the end, some must be in the middle, and some can be flexibly either at the beginning, or in the middle, or at the end. But why do we pay attention only to the pre circumstances and not to the post circumstances, for example, with the following notation:

> simple statement without noun clause as real subject:
>   pre circumstances,   simple SVerb phrase,  post circumstances.

We do so in order to avoid ambiguity of the circumstances. An example is the sentence with a noun clause as the object of the sentence. For example, for the sentence "*Today I know he will come tomorrow.*" the post circumstance "*tomorrow*" should be recognized as the circumstance of the object, i.e. "*he will come tomorrow.*" But with the notation above it can be recognized as the circumstance of the main sentence, what is not correct. In the situation that the ambiguity is inevitable, for example, for the sentence "*I know he will come today*." we will still treat the post circumstance as the circumstance of the object.

So we don't treat the post circumstance as the common element of a simple statement, but include it in the object or predicate which are at the sentence end .

"A simple SVerb phrase" is a simple statement with the subject at the beginning and a verb phrase describing the behavior or state of the subject. The voice can be either active or passive. We distinguish the special "there be" sentences from other sentences: we label it with the word "there" as the virtual subject of the sentence, which does not exist in other normal sentences, and the actual subject as the nominal predicate of the virtual subject.

> simple SVerb phrase:
>   "there", be(NUMB, PERS, TENSE), mid circumstances, subject(NUMB, PERS)/
>     <subject><noun><word>there</word></noun></subject><verb_phrase><verb_type>be</verb_type
> ><tense>", TENSE, "</tense><numb>", NUMB, "</numb><pers>", PERS, "</pers>", be, "<predicate>
> <predicate_type>np</predicate_type>", subject, "</predicate>", mid circumstances, "</verb_phrase>";
>           #e.g.  There is a book on the desk.
>   subject(NUMB, PERS), Verb phrase(NUMB, PERS)/
>       "<subject>", subject, "</subject><verb_phrase>", Verb phrase, "</verb_phrase>".

In the notation we label the subject with the tag pair <subject>...</subject> and the verb phrase with the pair <verb_phrase>...</verb_phrase>.

The subject is essentially a noun phrase. Attention should be paid to the correspondence or agreement between the subject and their verb phrase. The subject has the attribute of number (single or plural, described by the affix--NUMB:: sing | plur.) and personality (first, second, and third, described by the affix--PERS :: first | second | third.). The verb (lexical or auxiliary) in the verb phrase has also these attributes. Grammatically these attributes should be the same one. This character can be guaranteed by the AGFL notation.

### 2.2.4.3  Question

A question sentence is either a compound question, or a complex question, or a simple question. The max-matching mechanism is also used in parsing the question sentence.

A complex question is a main simple question plus a subordinate clause.

A compound question is only "and or question" which consists of simple questions connected by "and", "or", or comma.

So the question sentences can be decomposed into simple questions, and sometimes with simple statements.

### 2.2.4.4  Order

An order sentence is either a compound complex order, or a compound order, or a complex order, or a simple order. The max-matching mechanism is also used in parsing the order sentence.

A complex order can either be a subordinate clause plus a main simple order, or a main simple order plus a subordinate clause. Between the subordinate clause and the main clause there may be, and sometimes must be, a comma.

A compound order can be divided into two types: "and or order", and "two simple orders". The "two simple orders" is just two simple order sentences connected by such conjunctions as "but". In the "and or order" the last independent simple order is attached to other order sentences with the conjunction "and" or "or", whilst the others are connected by comma.

A Compound complex order is actually a subordinate sentence plus an "and or order" sentence. So the orders can be decomposed into simple orders, and sometimes with simple statements.

Summarily we can see that all kinds of statement sentences, question sentences and order sentences consist of simple statements, or simple questions, or simple orders. As the descriptions with NLML above show, the system label all the simple ones with the tag pair <simple_sentence>…</simple_sentence>, i.e. these simple ones (statement, question, order) can be treated as one type of sentence—simple sentence. Now we demonstrate that this method is correct.

### 2.2.4.5  Full Exclamation

There are two types of full exclamation sentences. If the emphasized part is the noun predicate, the "what" is put at the beginning of the sentence. If the emphasized part is the adjective predicate, the "how" is used.

## 2.3  Clauses

Noun clause and relative clause have at least a verb phrase as well as a subject which may be hidden in their main sentence in some cases, so they can be parsed as a simple statement sentence.

The noun clause can work as a noun phrase, for example, as the subject, object or complement in the sentence, or the object in the prepositional phrase. Many kinds of noun phrases should be considered: normal infinitive clause, query infinitive clause, negative infinitive clause, gerund clause, negative gerund clause, possessive gerund clause, that-clause, whether(if)-clause, query noun clause, etc.

The relative clause is located after the modified noun phrase. There are two forms of relative clause.  The full relative clause has a subject and at least one verb phrase, yet the subject, the complement or the object represented by a query noun must be located at the left of the phrase. The terse relative clause including infinitive clause, present participle and past participle, omits the subject of the clause.

## 2.4   Phrases

### 2.4.1   Verb Phrase

We should think about the complicated verb phrases with two or more verb parts connected by conjunctions. Here we use the tag pair <verb_phrase_part>…</verb_phrase_part> to include the "simple Verb phrase" without any conjunction, and the <verb_phrase_connector>…</verb_phrase_connector> to label the conjunction.  So the kernel in parsing the Verb phrase is the "simple Verb phrase(NUMB, PERS)" without any conjunction. The NUMB and PERS are the attributes of this phrase. The NUMB (number) can have the value of "sing"(singular) or "plur"(plural). The PERS can have the value of "first", "second" or "third".

The simple verb phrase can be active or passive. The simple active verb phrase is represented by the "all Verb phrase" with all kinds of possible tenses used in a possible sentence represented by the meta label SENTENCETENSE:present, past, present progressive, past progressive, and the modal verb plus the different form of the lexical verb, etc..

There are two types of all Verb phrase: real lexicon verb phrase, and be plus predicate phrase. We use still the max-matching mechanism to avoid the ambiguity if the "be" appears in the sentence and can be explained both as a real lexicon verb (plus predicate) and as an auxiliary verb preceding a progressive verb phrase. An example is the sentence "*I am doing the job.*" Grammatically this sentence can be explained in this way:

I--subject, am doing--verb(present continuous), job--object

Or:  I-subject, am--be(present), doing the job—gerund clause as a predicate.

Using the max-matching mechanism we can avoid the second parsing result.

The "real all Verb phrase" contains a lexicon verb and may contain different kinds (>30) of attachments to describe the behavior or the state of the subject, e.g. transitive verb + object, bitransitive verb + indirect object + direct object, verb + particle + prepositional  phrase, +verb + particle + object, verb + object + bare infinitive clause, verb + object  + past participle clause, etc. We use the tag pair
<verb_type>...</verb_type> to label the semantic type of the lexicon verb,
<direct_obejct>...</direct_object> to describe the direct object, the
<indirect_object>...</indirect_object> to describe the indirect object, the
<prep_phrase>...</prep_phrase> to describe the predicate, and the
<predicate>...</predicate> to describe the predicate, etc.

The simple passive verb phrase consists of the different form of "be" and the "all Verb phrase" with the tense "past participle".

### 2.4.2 Noun Phrase

Nouns are used to define the names of people, places and things. They describe the objects in the world or the subjective states in the human brain. This phrase has three attributes: NUMB(number), PERS(personality) and CASE. As in English the accusative form is the same as the dative one, there are only two values for this affix: nom (nominative) and dat (dative).

A noun phrase can consist of several parts connected by conjunctions. Here we use the tag pair <part>…</part> to include the "noun part" without any conjunction, and the <part_connector>…</part_connector> to label the conjunction. So the kernel in the Noun phrase is the "noun part(NUMB, PERS, CASE)" without any conjunction. It must have the kernel part (noun, pronoun, number, etc.) and may have the pre modifiers like article, demonstrative, adjective, etc. and the post modifiers like prepositional phrase and relative clause.

### 2.4.3  Adjective Phrase

There are two types of adjective phrases: attribute adjectives which are place before the modified noun phrase, and predicate adjectives which are used as the predicate of this sentence. The attribute adjectives consist of the pure attribute adjectives which can only be used as the attribute and the normal adjectives which can be used either as attribute or as predicate. The predicate adjectives consist of the pure predicate adjectives which can only be used as the predicate and the normal adjectives. A predicate adjective phrase can consist of several parts connected by conjunctions. We use the tag pair <part>…</part> to include the "predicate part" without any conjunction, and the <part_connector>… </part_connector> to label the conjunction.

Besides the normal adjective words the present participle and past participle can also be used as adjective phrases.

The adjective phrase has an affix GRAD. It can be absolute, comparative, superlative, or predicative. The GRAD of the attribute adjective may be one the first threes. The pure predicate adjective can only have the predicative GRAD. The normal attribute adjectives with the comparative GRAD can have "than" plus a noun phrase or a simple statement sentence as its compared object. The adjective phrase can also be used together with the conjunction such as "so...as", "as...as", "so...that", "too...to", or "enough...to", and the noun clause to express the result or the extend of the adjective meaning.

Some adjectives can have another attachment such as a prepositional phrase, or infinitive clause, or other noun clause. The adjective phrase can have preceding adverb words which modify the adjective phrase.

### 2.4.4.  Adverb Phrase

The adverb phrase consists of the pure adverb words which have no affix of GRAD and some special adjectives with only one syllable which can also function as adverbs and have the affix GRAD. The pure adverb can have the preceding word "more" to express the comparable state, and the "the most" to express the superlative state.

### 2.4.5   Prepositional Phrase

The prepositional phrase is made up of a preposition and its object. The preposition can be one word, or a fixed expression with several words. The object can be a noun phrase, which may have relative clause modifying it, or a noun clause (gerund clause or query infinitive clause).

### 2.4.6   Circumstance Phrase

The circumstances can be put at the beginning of a sentence (pre circumstances), in the mid (mid circumstances), or at the end (post circumstances).

The pre circumstances can either be a mid circumstance or a participle phrase (past or present).

The mid circumstance can either be the adverbs expressing the time, place, way, etc. or prepositional phrase.

The post circumstance can be either a mid circumstance or a participle phrase (past or present). If the kernel is an adverb word it may be used together with the conjunction such as "so...as", "as...as", "so...that", "too...to", or "enough...to", and the noun clause to express the result or the extent of the adverb word. If the kernel is an adverb in the comparative form, it can have "than" plus a noun phrase or a simple statement sentence as its compared object.

### 2.4.7   Predicate Phrase

The predicate phrase may be a noun phrase with the nominative case, or a noun clause, or a preposition phrase, or an adjective phrase. We use the tag pair <predicate_type>…</predicate_type> to distinguish the different kinds of predicates and to parse then correspondently.

## 2.5   Words

The phrase is made of up different kinds of lexical words: adjectives, adverbs, nouns, pronouns, verbs, etc. All the basic forms of the lexical words are included in the tag pair <word>...</word>. Every type of the words together with their possible affix and statistical probability[1], are stored in a lexicon data file, which is an easily editable text file.

## 2.6 One Example of NLML

The NLML for the sentence "*If it rains today, you will not go, and I will not come*" is:
<mood>statement</mood><complexity>compound complex</complexity><subordinator> if </subordinator><sub><subject><noun><type>perspronoun</type><word>it </word><numb>sing</numb><pers>third</pers><case>nom</case></noun></subject><verb_phrase><verb_type>verb</verb_type><tense>present</tense><numb>sing</numb><pers>third</pers><verb_word>rains </verb_word><circum></circum><circum><circum_type>adv</circum_type><adv><type>time</type><word>today</word></adv></circum></verb_phrase></sub>, <complete_sentence><subject> <noun><type>perspronoun </type>

---

[1] By parsing the system will consider this probability distribution. This is called hybrid parsing according to the AGFL notation.

<word>you</word><numb>NUMB</numb><pers>secnd</pers><case>nom</case></noun></subject><verb_phrase><verb_type>verb</verb_type><tense>modal</tense><numb>NUMB</numb><pers>second</pers><verb_word>will not </verb_word><verb_word>go</verb_word><kernel_tense>infi</kernel_tense></verb_phrase></complete_sentence><complete_sentence><subject><noun><type>perspronoun</type><word>I </word><numb>sing</numb><pers>first</pers><case>nom</case></noun></subject> <verb_phrase><verb_type>verb</verb_type><tense>modal</tense><numb>sing</numb><pers>first</pers><verb_word>will not </verb_word><verb_word>come </verb_word><kernel_tense>infi</kernel_tense> </verb_phrase></complete_sentence><sentence_connector>and</sentence_connector>

# 3    Applications of NLML in NLP

The NLML can be widely used in NLP-programs, for example, in information retrieval, in discourse storage and analysis, in natural language understanding and generating. We have developed NLML in the scope of a human-computer-interaction program for foreign language learning [10], so we introduce here two applications of the NLML in our project: NLOMJ and NLDB.

## 3.1 NLOMJ

As we have mentioned above, the grammar elements in every level of the natural language (sentences, clauses and phrases) have some common feathers as well as their specialty and are associated with each other. The two characters can be reflected by the concept of inheritance and association in the OOP (Object Oriented Programming). So we have selected Java, the typical OOP language to represent the grammar elements. We call this method NLOMJ (Natural Language Object Modal in Java).The UML diagram for the NLOMJ is shown in Fig.2.

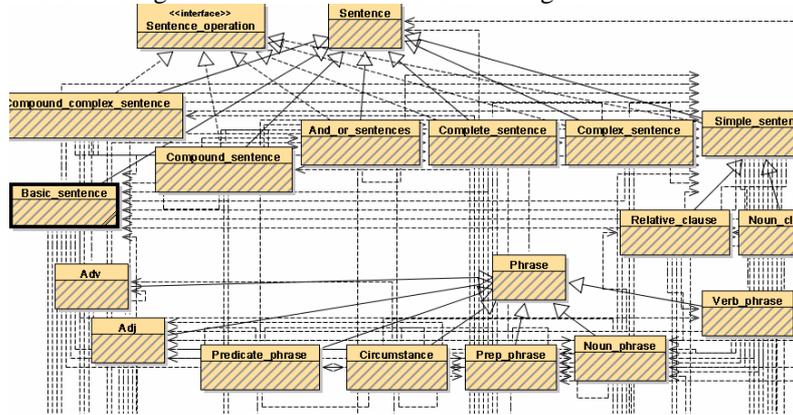

**Fig. 2.** The UML diagram for NLOMJ

With the objects of the grammar elements and relations in the memory, we could say that the computer understands the English grammar, and it can run some operations on a given NLML through the easy manipulation of markup language, e.g. negation, mood transformation, etc.

### 3.2 NLDB

With NLOMJ we can build the objects map of the syntax and semantics of the natural language from NLML. In this sense we say the NLML is a symbol form of the syntax and semantics. For the long-term usage of the syntax and semantics we can save the NLML as a String directly in the table of a database. We can classify the NLML according to our requirements, for example, into facts (simple statement), question and the relations of the facts (complex statement with subordinate clause), and then save them in different tables of the database. After retrieving the NLML from the database we can rebuild the objects map of the historical text or discourse into the active memory for the further analysis. We call this technique NLDB (Natural Language Database). As for our CSIEC-Project the NLDB constructs the basis of the context analysis for the dialog generating.